\documentclass[10pt,twocolumn,letterpaper]{article}

\usepackage{cvpr}              % To produce the CAMERA-READY version
\usepackage{subcaption}
\usepackage{multirow}
\usepackage{booktabs}
\usepackage{makecell}
\usepackage{natbib}
\usepackage{graphicx}
\usepackage{float}                  % 图片浮动位置           
\usepackage{overpic}  
\definecolor{cvprblue}{rgb}{0.21,0.49,0.74}
\usepackage[pagebackref,breaklinks,colorlinks,allcolors=cvprblue]{hyperref}

\title{Efficient Bayer-Domain Video Computer Vision with Fast Motion Estimation and Learned Perception Residual}

\author{Haichao Wang\\
SIGS, Tsinghua University\\
Shenzhen, China\\
{\tt\small hychaowang@outlook.com}
\and
Jiangtao Wen\\
New York University\\
New York, USA\\
{\tt\small jw9263@nyu.edu}
\and
Yuxing Han\\
SIGS, Tsinghua University\\
Shenzhen, China\\
{\tt\small yuxinghan@sz.tsinghua.edu.cn}
}

\begin{document}
\maketitle
\begin{abstract}
Video computer vision systems face substantial computational burdens arising from two fundamental challenges:  eliminating unnecessary processing and reducing temporal redundancy in back-end inference while maintaining accuracy with minimal extra computation.
To address these issues, we propose an efficient video computer vision framework that jointly optimizes both the front end and back end of the pipeline. On the front end, we remove the traditional image signal processor (ISP) and feed Bayer raw measurements directly into Bayer-domain vision models, avoiding costly human-oriented ISP operations. On the back end, we introduce a fast and highly parallel motion estimation algorithm that extracts inter-frame temporal correspondence to avoid redundant computation. To mitigate artifacts caused by motion inaccuracies, we further employ lightweight perception residual networks that directly learn perception-level residuals and refine the propagated features.
Experiments across multiple models and tasks demonstrate that our system achieves substantial acceleration with only minor performance degradation.
\end{abstract}    
\section{Introduction}
\label{sec:intro}

Recent years have witnessed rapid progress in video perception applications, benefiting from great success of computer vision, with significantly successful models including CNN-based and transformer-based vision models~\cite{he2016deep, dosovitskiy2020image, sandler2018mobilenetv2, liu2021swin, yu2018bisenet, long2015fully, bochkovskii2024depth}. As these algorithms increasingly move to real-world applications like mobile and embedded platforms, efficiency and latency have become critical constraints.

A substantial portion of the computational budget on mobile and edge devices is consumed not by the vision model itself, but by the camera’s image signal processor (ISP)~\cite{qian2022rethinking, baek2008noise, nishimura2018automatic, afifi2020deep}. Typical pipelines introduce 1–5 ms processing delay and can exceed 300–600 mW under high-resolution or multi-frame workloads. The ISP performs demosaicing, denoising, and color correction to produce human-readable images before they are passed to downstream models. However, these operations are designed for human eyes rather than machine vision, introducing redundant processing and nontrivial latency for video computer vision systems.
Meanwhile, on the back end, video computer vision models remain inefficient when executed in a frame-by-frame manner. High frame rates lead to significant temporal redundancy, causing models to repeat similar computations for consecutive frames and ultimately wasting inference resources. 

% The problems mainly come from two aspects. On the front end, there exists unnecessary and complex processing on raw data. In traditional video computer vision systems, photons are captured by the sensor and then filtered by the Bayer array, outputting the Bayer raw data. After that, the Bayer Raw data goes through a series of image signal processing operations, which are computationally complex and with low energy efficiency. These processes create the human-readable RGB images, which are then sent into computer vision models. However, for computer vision models, the transition from Bayer raw data into RGB images is unnecessary, because computer vision models can directly process the Bayer raw data.

\begin{figure}
    \centering
    \includegraphics[width=1\linewidth]{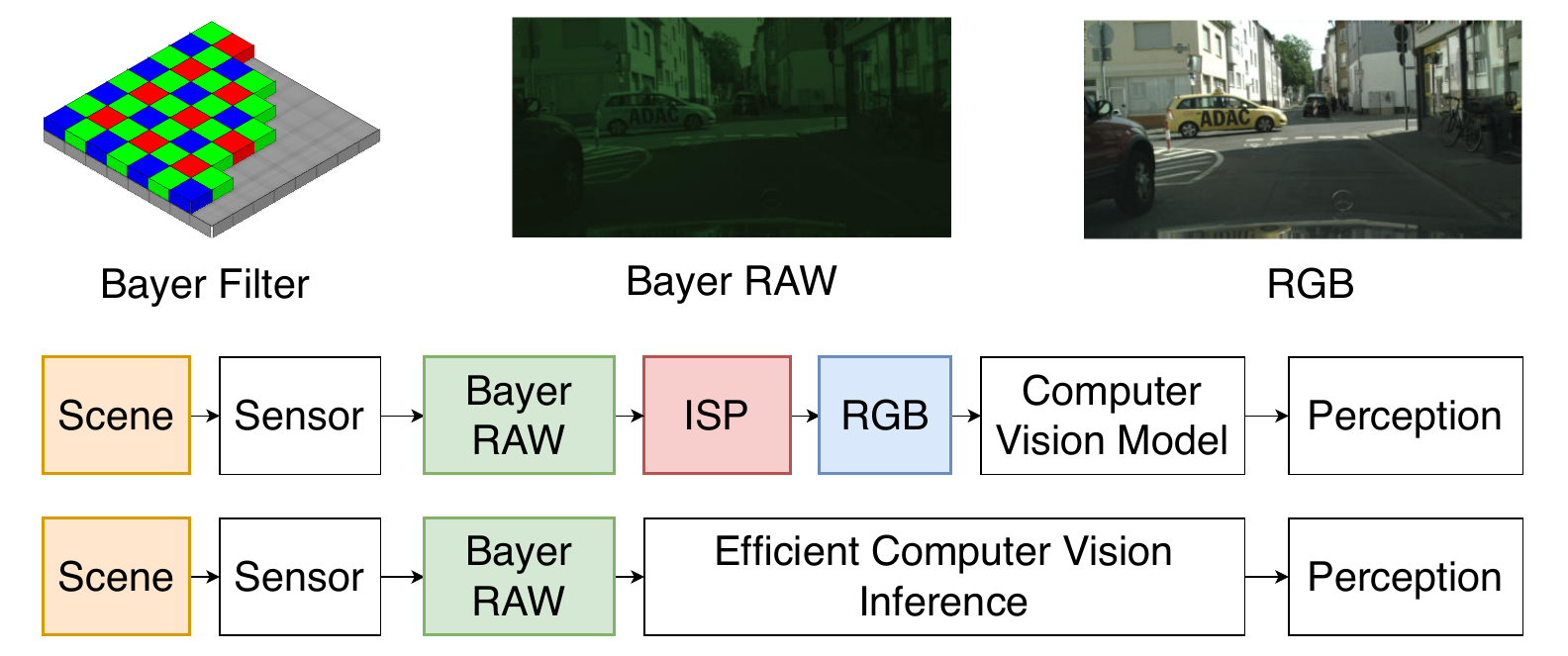}
    \caption{Top: Traditional pipeline transforming Bayer raw data to RGB data with ISP. Bottom: Our proposed non-ISP pipeline, directly processing Bayer raw data.}
    \label{fig:pipeline}
\end{figure}

Several approaches~\cite{kondratyuk2021movinets, li2019dfanet, wang2025efficient, jain_accel_2019, hu_efficient_2023} improve video inference efficiency by leveraging the classic motion estimation and residual map correction from video coding. They propagate predictions from a reference frame using motion vectors and then refine the warped results with the guidance of residual maps. For instance, TapLab~\cite{feng_taplab_2022} identifies regions with large residual responses and selectively recomputes them.
Despite their effectiveness, these approaches face notable limitations. They rely either on codec-derived motion vectors~\cite{wiegand2003overview, pastuszak2015algorithm}, which are coarse and CPU-bound, or on deep optical-flow models~\cite{dosovitskiy2015flownet, ilg2017flownet, hui2018liteflownet, sun2018pwc}, which are too expensive for real-time use. Their refinement is further guided by residual maps, whose magnitude poorly reflects perceptual or semantic correctness, often causing inaccurate correction. Moreover, these methods focus solely on back-end efficiency and overlook front-end ISP computation, which remains a major bottleneck in practical systems. 

\begin{figure}
    \centering
    \includegraphics[width=1\linewidth]{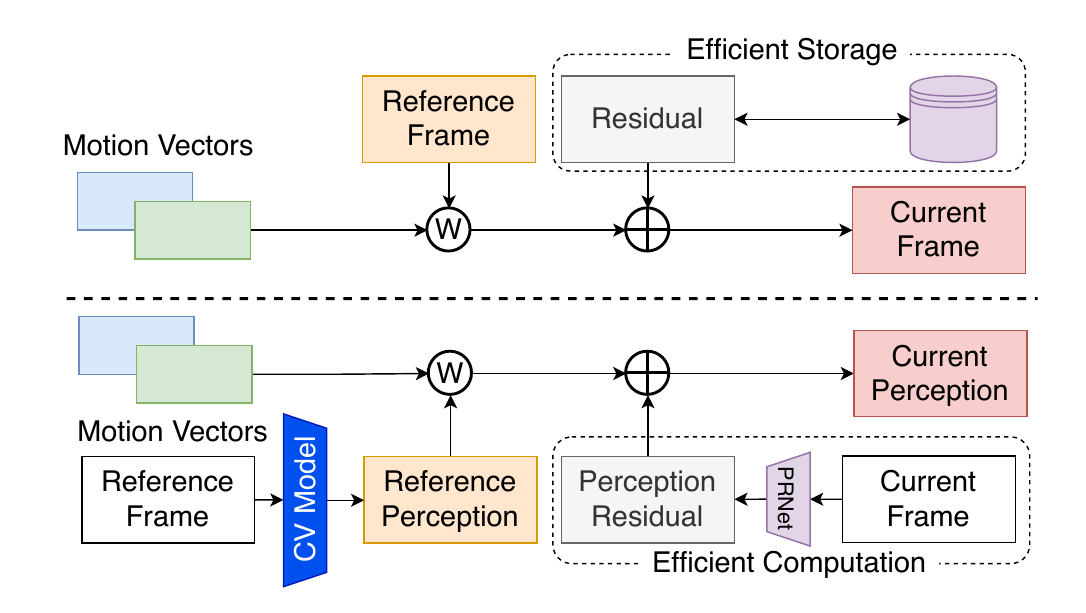}
    \caption{Top: Image residual to correct the current frame. Bottom: Perception residual to correct computer vision perception.}
    \label{fig:perception_residual}
\end{figure}

To build an efficient video computer vision system, the key challenges are:
1) how to eliminate unnecessary computation from the video computer vision pipeline, and
2) how to effectively reduce temporal redundancy while preserving accuracy with minimal additional computational overhead.
To solve these problems, we propose an efficient video computer vision system that jointly optimizes both the front end and back end of the pipeline, as illustrated in Figure~\ref{fig:pipeline}. On the front end, we remove the image signal processor entirely and feed Bayer raw data directly into models trained in the Bayer datasets converted by Invertible-ISP~\cite{xing2021invertible}, eliminating expensive ISP operations and reducing latency. On the back end, we introduce a new framework for efficient video inference. We first develop a fast motion estimation module with a pyramid block structure and coarse-to-fine search, designed for highly parallel execution on modern GPUs.
To enable accurate refinement, we extend the classical residual concept from the image to the perception, as illustrated in Figure \ref{fig:perception_residual}. Rather than relying on residual-map magnitude for guidance, we use lightweight models to directly learn perception-level residuals, providing stable and effective correction numerically. In conclusion, our main contributions are summarized as follows:
\begin{enumerate}
\item We remove the ISP from the pipeline and process Bayer raw data directly with Bayer-domain models, bypassing slow and complex image signal processing operations.
\item We propose a fast and parallel motion estimation module that significantly reduces the computational cost and delay.
\item We extend the residual concept to perception features and correct predictions using learned perception residuals.
\item Extensive experiments across multiple models and tasks demonstrate that our system achieves substantial acceleration with only minor performance degradation.
\end{enumerate}

\section{Related Works}
\label{sec:formatting}
\begin{figure*}
    \centering
    \includegraphics[width=1\linewidth]{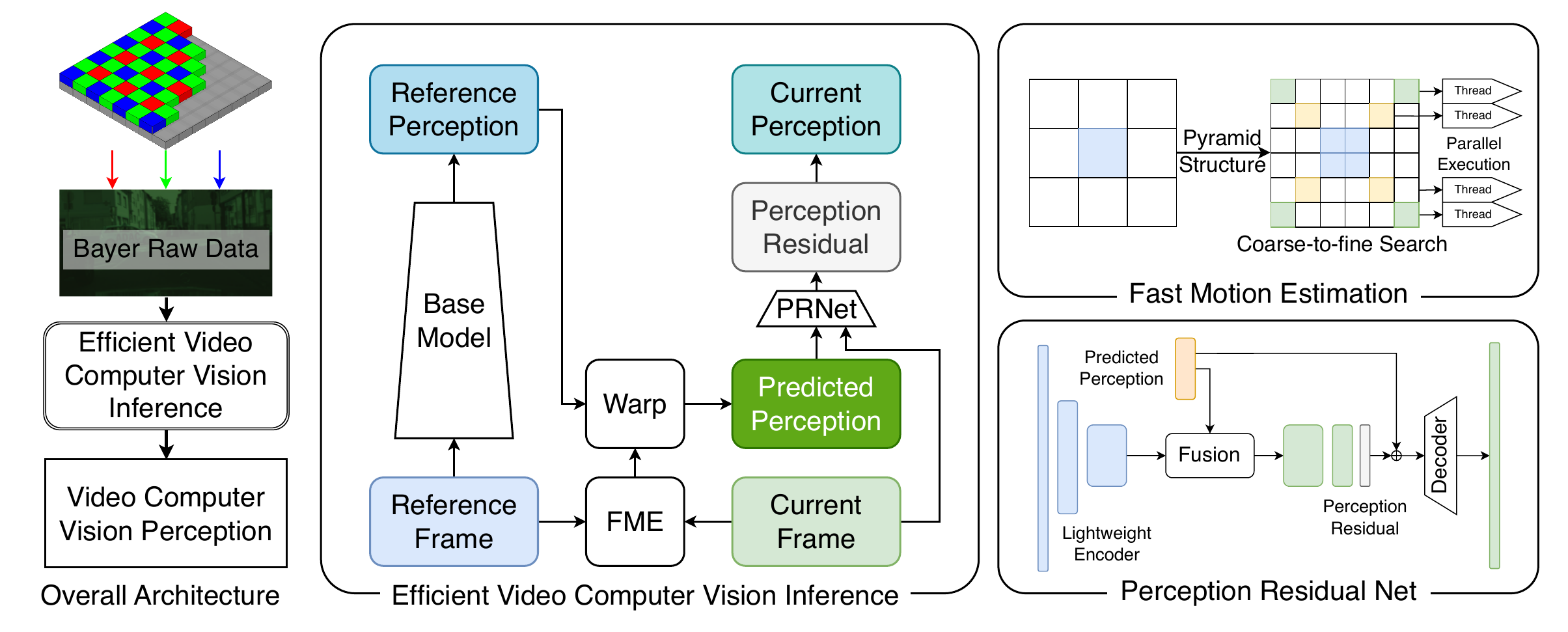}
    \caption{Overall Architecture of our proposed efficient Bayer-domain video computer vision system.}
    \label{fig:main}
\end{figure*}
\subsection{Video Computer Vision Acceleration}
Improving the efficiency of video computer vision has been a long-standing challenge. Early works explicitly exploit inter-frame redundancy to reduce computation. Clockwork~\cite{shelhamer2016clockwork} performs sparse inference via keyframe segmentation, while Mahasseni et al.~\cite{mahasseni2017budget} interpolate predictions from neighboring frames. To mitigate information loss in non-key frames, Li et al.~\cite{li2018low} fuse shallow features extracted from the current frame with propagated deep features using spatially variant convolution.

Subsequent methods leverage motion estimation to further reduce temporal redundancy. Optical-flow-based propagation, such as in Zhu et al.~\cite{zhu2017deep} and Xu et al.~\cite{xu2018dynamic}, warps key frame features to generate non-key frame predictions. However, inaccurate motion often introduces propagation errors. To address this, TapLab~\cite{feng_taplab_2022} incorporates codec-derived motion vectors together with residual maps to selectively recompute unreliable regions. Xiong et al.~\cite{xiong2021distortion} additionally exploit edge cues and residual signals to refine warped features.
Despite these advances, existing approaches still struggle with inefficient redundancy detection, either relying on coarse inter-frame differences or computationally heavy flow estimation, and inaccurate residual-based correction, caused by the mismatch between pixel similarity and prediction similarity.

Beyond explicit redundancy modeling, several studies improve efficiency through feature fusion across frames. TD-Net~\cite{hu2020temporally} aggregates multi-timestamp features and distributes inference across multiple lightweight models. AR-Seg~\cite{hu_efficient_2023} enhances non-key frame predictions by using high-resolution key frame features for rectification. Although reducing per-frame costs, they either neglect the motion information or overlook the computation redundancy when facing little changes.

\subsection{Motion Estimation and Compensation}
\noindent \textbf{Definition} Motion Estimation is an analytical process to find the optimal Motion Vector ($\text{MV}$) by minimizing a distortion metric between the current frame $F_C$ and reference frame $F_R$. This process is expressed as $\text{MV}=\text{ME}(F_R, F_C)$. Motion Compensation (MC) is a predictive process that uses the MV to generate a prediction. It transforms (\text{Warp}) the reference frame $F_R$ using the MV and computes the prediction Residual ($\text{Res}$).The process can be represented as $F_C=\text{Warp}(F_R, \text{MV})+\text{Res}$, where $\text{Res}$ works as a numerical correction for the predicted frame.

\noindent \textbf{Block Matching-based ME.} Video codec tools predominantly bases on Block-Matching Algorithms. For instance, the Full Search algorithm provides the optimal match but suffers from extremely high computational complexity. To achieve real-time encoding, various fast-search algorithms have been proposed, such as Diamond Search and Hexagonal Search, which significantly reduce computational load by sacrificing minor accuracy.

\noindent \textbf{Optical Flow-based ME.} Optical flow methods can work as ME algorithms, which starts with gradient-based modeling like Lucas-Kanade algorithm and Horn-Schunck algorithm. In recent years, deep learning-based methods have been introduced into optical flow estimation. Convolutional Neural Network models, represented by FlowNet~\cite{ilg2017flownet}, PWC-Net~\cite{sun2018pwc}, and RAFT~\cite{teed2020raft}, have achieved state-of-the-art performance in both accuracy and speed by training end-to-end on large synthetic datasets, surpassing traditional approaches.

ME methods mentioned above usually focus on video compression, having complex design and high computation costs. Block matching-based methods usually contains complex serial operations, suitable for CPU. Optical flow methods have intensive computation, especially with deep learning models. In this paper, we propose an ME method, both concise in algorithm design and parallel in implementation, for video computer vision acceleration.
\section{Methods}
\subsection{Overall Architecture}

As illustrated in Figure~\ref{fig:main}, we propose an efficient computer vision pipeline that eliminates the conventional ISP and feeds raw sensor measurements directly into an efficient video computer vision back end.

To support this design, we construct a synthesized Bayer-format dataset by applying a novel Invertible-ISP model to large-scale RGB datasets, effectively reversing the ISP process. Standard computer vision architectures are adapted by modifying their input layers to accept single-channel Bayer patterns and are then retrained on this transformed data. During inference, the back-end model operates directly on sensor-captured Bayer frames, removing ISP-induced latency and computational overhead.

To achieve efficient perception on Bayer-domain inputs, we introduce a lightweight video inference framework at the back end of the pipeline. A Bayer-domain base model first processes a key frame to obtain a high-quality reference prediction. For subsequent frames, motion vectors are estimated using the fast motion estimation method described in Section~3.2, capturing temporal correspondence between the reference and current frames. These motion vectors are then used to propagate the reference perception, producing coarse predictions for intermediate frames. Finally, perception residual net (PRNet) described in Section~3.3, trained to estimate perception-level residuals, corrects the propagated features to generate accurate and stable final outputs.

\subsection{Fast Motion Estimation}

As discussed earlier, motion estimation (ME) is a key component of our overall framework. However, existing ME approaches—either by directly invoking video codec tools or by applying deep optical flow models—fail to satisfy real-time requirements due to their algorithmic complexity and limited execution efficiency. Codec-derived motion vectors involve substantial and unnecessary overhead, such as coding-tree traversal and test-zone evaluation, and are typically executed serially on CPUs, resulting in low throughput. Optical-flow–based methods, on the other hand, rely on deep neural networks and remain computationally expensive even when accelerated on GPUs.

To overcome these limitations, we design a lightweight, fully GPU-executable motion estimation algorithm composed of three components: a pyramid block structure, a parallel coarse-to-fine block matching scheme, and an efficient matching criterion.

\begin{figure}
    \centering
    \includegraphics[width=1\linewidth]{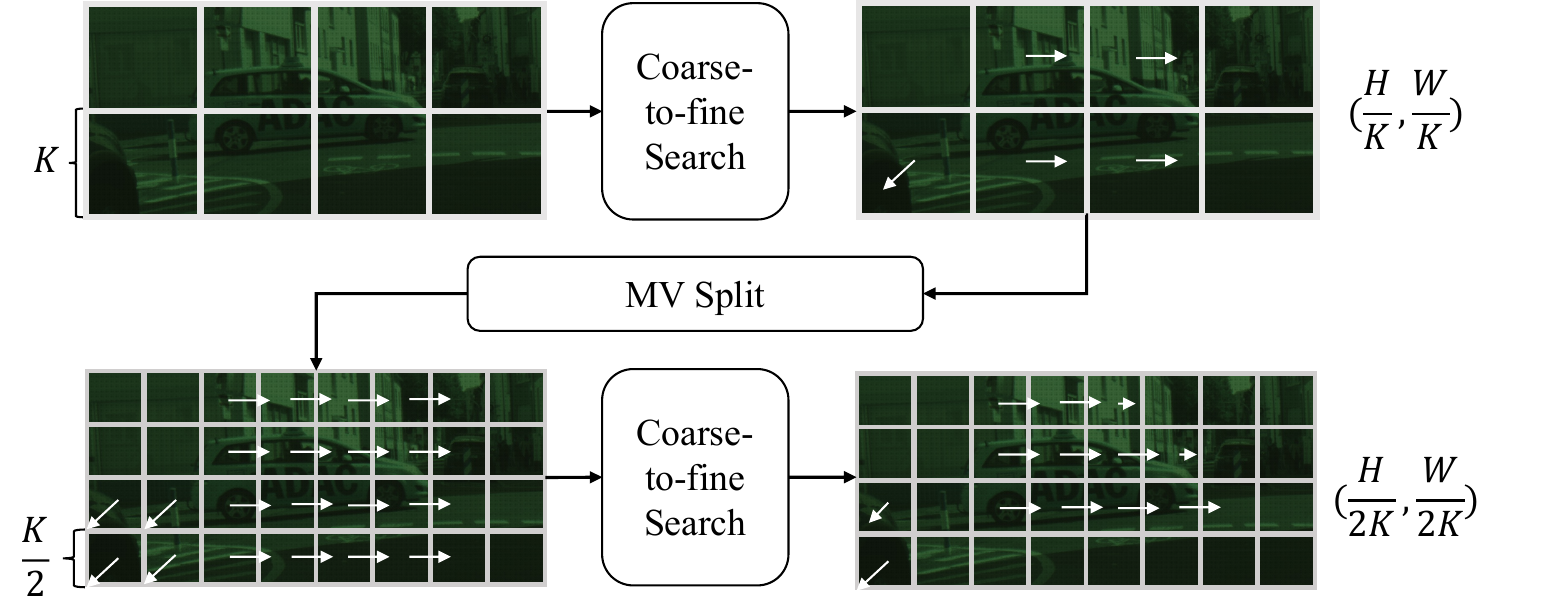}
    \caption{The pyramid structure of motion estimation.}
    \label{fig:pyramid}
\end{figure}

\noindent \textbf{Pyramid Block Structure.} To achieve accurate perception propagation, the ME module must reliably match blocks with similar visual structures. While pixel-level similarity often correlates with feature-level similarity, this assumption holds more consistently when blocks contain sufficient contextual information—i.e., when using larger block sizes. Conversely, smaller blocks are necessary to capture fine-grained motion.
We therefore adopt a pyramid block architecture that integrates both advantages. Large-block matching is first performed to provide robust global alignment, followed by progressively smaller blocks to refine local correspondence. Unlike traditional motion estimation designed for compression—where minimal block sizes of 4 or 8 pixels are used—our approach employs larger minimum block sizes (16 or 32 pixels) to ensure a better balance between computational efficiency and matching accuracy.

It is important to note that our block pyramid differs fundamentally from conventional resolution pyramids. Resolution pyramids downsample feature maps to enlarge the search range, but this downsampling weakens pixel-level constraints and often leads to inaccurate alignment. In contrast, our block pyramid preserves full-resolution content while varying only the block size, thereby maintaining strong pixel–feature consistency across scales.

\noindent \textbf{Parallel Coarse-to-fine Block Matching.}
To balance accuracy and speed, block matching is performed through three stages: coarse, intermediate, and fine. The coarse stage uses a large search range and step size to rapidly identify candidate motion regions; subsequent stages progressively narrow both to refine the estimate.

\begin{figure}
    \centering
    \includegraphics[width=\linewidth]{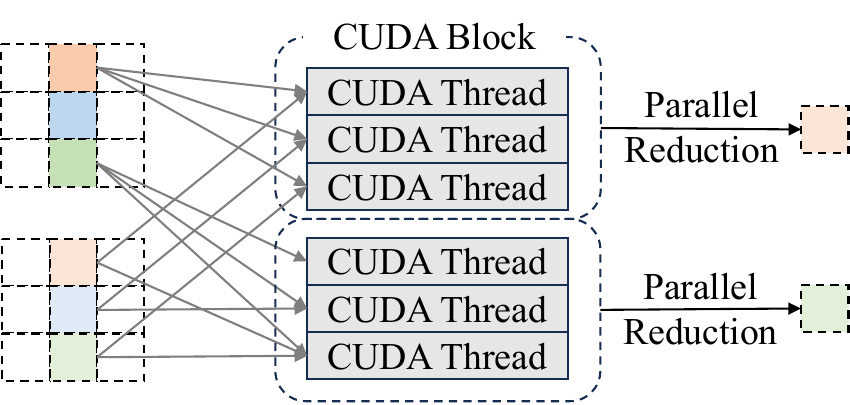}
    \caption{The diagram of fast motion estimation for video computer vision. Each block pair is processed in a CUDA thread for parallel.}
    \label{fig:FPME}
\end{figure}

The major computational burden in ME arises from evaluating block similarity across the search range. To make this process highly efficient, we design a GPU-parallel implementation. In each matching stage, every block in the current frame is assigned to a dedicated CUDA block whose threads evaluate candidate matches in parallel, as illustrated in Figure~\ref{fig:FPME}. Each CUDA block loads the target block into shared memory to reduce global memory access and maximize data reuse. This design yields a highly efficient parallel block-matching procedure tailored to modern GPU architectures.

\noindent \textbf{Matching Criterion.} We adopt the Sum of Absolute Differences (SAD) as the similarity metric due to its simplicity and hardware efficiency. The candidate block with the lowest SAD is selected as the best match, and the offset between matched block centers is recorded as the motion vector (MV).

In summary, our ME algorithm provides a fast, accurate, and GPU-optimized motion estimation framework that maintains strong consistency between pixel-level structures and high-level perception features, enabling efficient propagation across video frames.

\subsection{Video Perception Residual}

For reconstruction of frames, in video compression, a numerical difference between the predicted frame and target frame is recorded, which is then used to correct the prediction during decoding. The numerical difference, called residual map, saves the storage because of its sparsity.

Different from traditional method, using residual map to find the correction region, we extend the residual concept into perception-level. In this section, we propose video perception residual, a numerical correction for predicted video perception and re-formulate the correction process to:
\begin{align}
    \textbf{z}_C = \text{Warp}(\textbf{z}_R, \textbf{MV}) + \textbf{z}_\text{Res} \nonumber
\end{align}
where $\textbf{z}_\text{Res}$ denotes the video perception residual. Because of the temporal correlation between frames, similar to pixel-level residual map, the perception residual map is sparse, which means using a lightweight model can learn, where PRNet is proposed. 

PRNet is lightweight in design, with different scales for different level of video perception residual. The network begins with a compact encoder built upon depth-wise separable convolutions, which extracts low-level structural cues from the current Bayer-frame input through stride-2 downsampling. These features retain fine textures and edge information essential for correcting the warped prediction. PRNet then fuses the low-level feature map with the propagated feature from the reference frame by concatenating them and applying a 1×1 convolutional projection, forming a joint representation that captures both coarse perceptual predictions and frame-specific details. The fused feature is refined through a sequence of inverted residual blocks, which employ channel expansion, depth-wise filtering, and channel compression to achieve strong representational capability while maintaining minimal computational cost. A final $1\times1$ convolution predicts the perception residual, which is added to the propagated feature to obtain the refined output.

PRNet learns the video perception residual, using $\textbf{z}_C-\textbf{z}_\text{pred}$ as target, where $\textbf{z}_\text{pred}=\text{Warp}(\textbf{z}_R, \textbf{MV})$. For supervision, we adopt the Reversed Huber (BerHu) loss, which aligns with the goal of emphasizing large perceptual discrepancies. BerHu penalizes large errors more aggressively by switching to an L2 penalty when the residual exceeds a threshold. Since large perception mismatches have a stronger impact on downstream tasks, this loss formulation encourages PRNet to focus its capacity on the most critical corrections. In addition, to mitigate the imbalance between correctly and incorrectly predicted regions, we introduce region-size–based weighting. This prevents the network from converging to trivial solutions, such as simply predicting zero residuals everywhere, and encourages meaningful video perception residual learning.
\section{Experiments}

\subsection{Setup}

\noindent \textbf{Datasets and Tasks.}
We evaluate the proposed method on multiple datasets across two tasks: video semantic segmentation and video object detection.

For video semantic segmentation, we adopt two widely used benchmarks: CamVid and Cityscapes. The \textbf{CamVid} dataset~\cite{BrostowSFC:ECCV08, BrostowFC:PRL2008} contains 701 densely annotated frames sampled from 30 fps video sequences at a resolution of $720 \times 960$. Meanwhile, \textbf{Cityscapes}~\cite{cordts2016cityscapes} provides a more challenging large-scale benchmark featuring high-resolution frames ($1024 \times 2048$) captured from 17 fps videos. 
For video object detection, we employ the \textbf{ImageNet VID} dataset~\cite{ILSVRC15}, a large-scale benchmark for assessing temporal detection performance. Videos of varying resolutions are center-cropped to $512 \times 512$ to ensure consistent training and evaluation.

\noindent \textbf{Base Models.}
To demonstrate the compatibility and generality of our acceleration framework for video semantic segmentation, we select two representative image segmentation models: PSPNet\cite{zhao2017pyramid} and Segformer\cite{xie2021segformer}. PSPNet is a classical CNN-based model commonly used in prior work, enabling fair comparison. Segformer is a state-of-the-art transformer-based model with stronger baseline performance. For video object detection, we adopt DETR~\cite{carion2020end} as the base model. All base models are modified to accept single-channel Bayer raw inputs by adjusting the first convolution layer accordingly.

\noindent \textbf{Pre-processing.}
Since we remove the ISP from the image formation pipeline and feed raw signals directly into computer vision models, all base models are trained on Bayer raw data. We construct large-scale Bayer raw datasets from existing RGB datasets using Invertible-ISP~\cite{xing2021invertible}, which reverses the ISP pipeline to reconstruct Bayer raw signals. Dataset statistics (mean and variance) for normalization are recomputed on the reconstructed raw data.

\noindent \textbf{Training Details.}
Bayer models are trained with the same hyperparameters as their RGB counterparts. For training FRNet, we use the AdamW optimizer with an initial learning rate of $1\times10^{-4}$ across all datasets. The batch size is set to 16, and training is conducted for 100 epochs. A cosine annealing schedule is applied to gradually decay the learning rate to $1\times10^{-6}$ in the final epoch. All experiments are implemented in PyTorch 2.7.0 with CUDA 11.8, and training/testing are performed on a single NVIDIA A100 GPU. Additional model details are provided in the Appendix.

\noindent \textbf{Test Protocol and Metrics.}
Due to the sparse annotations in most video datasets, we evaluate the model using groups of pictures (GOPs). For a GOP length $L$, the model processes a frame sequence $[F_{i-L+1},\dots,F_i]$, with ground truth available only for the final frame $F_i$. The performance on $F_i$ is taken as the performance under GOP length $L$, aggregated across all sequences. We report mIoU for video semantic segmentation and mAP for video object detection. Computational cost is measured using FLOPs and runtime.

\noindent \textbf{CUDA Settings.}
For our FME implementation, each CUDA block is configured with $(R\times 2 + 1)^2$ threads, where $R$ denotes the search radius. Each CUDA grid is composed of (Batch, $h$, $w$) blocks, where $h$ and $w$ correspond to the height and width of the motion-vector field. Detailed CUDA settings are provided in the Appendix.

\subsection{Video Semantic Segmentation}
\noindent \textbf{Overall Results.} As shown in Table~\ref{tab:results}, with our framework, PSPNet-18 and Segformer-B5 are greatly accelerated on the video semantic segmentation task. 

\begin{table}[htbp]
    \centering
    \caption{Overall results on both datasets and models on the video semantic segmentation task.}
    \begin{tabular}{cccc}
    \toprule
        Model & Dataset & mIoU & GFLOPs\\
        \midrule
        PSPNet-18 & CamVid & 68.53 & 34.51\\
        PSPNet-18 & Cityscapes & 68.82 & 80.44 \\
        Segformer-B5 & CamVid & 74.98 & 30.78\\
        Segformer-B5 & Cityscapes & 79.81 & 67.50\\
        \bottomrule
    \end{tabular}
    
    \label{tab:results}
\end{table}

\noindent \textbf{Bayer-Domain Training.} We train computer vision models both on RGB and Bayer domain. The performance of the models are shown in Table~\ref{tab:bayertraining}, which shows similar performance in mIoU.

\begin{table}[htbp]
    \centering
    \caption{Results of our framework deployed on PSPNet-18 and Segformer-B5, with the standard setting of FME.}
    \label{tab:bayertraining}
    \begin{tabular}{lcccc}
        \toprule
        \multirow{2}{*}{Method} & \multicolumn{2}{c}{CamVid} & \multicolumn{2}{c}{Cityscapes} \\
        \cmidrule(lr){2-3} \cmidrule(lr){4-5} % 使用 cmidrule 替代 cline，(lr)表示左右留出空隙
        & RGB & Bayer & RGB & Bayer \\
        \midrule
        PSPNet-18   & 69.36\% & 68.89\% & 69.0\% & 69.35\%\\
        Segformer-B5 & 76.58\% & 75.78\% & 81.07\% & 80.02\% \\
        \bottomrule
    \end{tabular}
\end{table}

\noindent \textbf{Comparison.} Several previous works propagating features or results from reference frames are compared with our proposed method, with same testing protocols, base models and metrics. As shown in Table~\ref{tab:performace}, our method achieves the best performance-acceleration balance. Methods like TD, despite the slight increase in mIoU, suffer from a huge increase in computation. TapLab is a representative of over-correction, using a $512 \times 512$ window to re-calculation, even with a slight error. AR-Seg and Accel cost too much computation on the network for all frames with different residual scales. Jian et al. use the features from previous and next key frame, partly solving the artifacts, but depending on the key frames. BlockCopy neglects the motion information, leading to more error to correct. The success of our proposed method comes from the extremely fast motion estimation with concise algorithm and parallel implementation, as well as lightweight feature residual models for correction.

Moreover, it is worth mentioning that, we consider the computation of the whole back end processing. Motion estimation computation from video codec tools in methods like AR-Seg and TapLab is not included, which means their acceleration is over-estimated. Typically, runtime of H.264 video codec tools in ffmpeg ranges from 15ms to 50ms per frame, while H.265 with x265 library can reach 200ms.

\begin{table}[!htbp]
\centering
\caption{Comparison with previous works on PSPNet-18.}
\label{tab:performace}
\resizebox{\columnwidth}{!}{
\begin{tabular}{llccc}
\toprule
\multicolumn{2}{l}{\textbf{Method}} & \textbf{mIoU(\%)} & \textbf{GFLOPs} & \textbf{\begin{tabular}[c]{@{}c@{}}$\Delta$GFLOPs\end{tabular}} \\
\midrule
\multirow{9}{*}{\rotatebox[origin=c]{90}{CamVid}} & Base Model & 69.36 & 309.02 & - \\
& Accel & 66.15 & 397.70 & +28.70\% \\
& TD & 70.13 & 363.70 & +17.7\% \\
& BlockCopy & 66.75 & 107.52 & -45.7\% \\
& TapLab & 67.57 & 117.73 & -50.2\% \\
& Jain et al. & 67.61 & 146.97 & -53.8\% \\
& AR-Seg & 70.82 & 133.09 & -57.0\% \\
\cmidrule(lr){2-5} % 使用 \cmidrule 替代 \cline
& Bayer Model & 68.89 & 308.19 & - \\
& Ours & 68.53 & 34.51 & -88.45\% \\
\midrule % 使用 \midrule 分隔两个数据集，替代 \hline \hline
\multirow{9}{*}{\rotatebox[origin=c]{90}{Cityscapes}} & Base Model & 69.00 & 560.97 & - \\
& Accel & 68.25 & 1011.75 & +96.0\% \\
& TD & 70.11 & 673.06 & +20.0\% \\
& BlockCopy & 67.69 & 294.20 & -41.2\% \\
& TapLab & 68.90 & 237.29 & -50.6\% \\
& Jain et al. & 68.57 & 342.67 & -52.5\% \\
& AR-Seg & 69.45 & 234.91 & -58.1\% \\
\cmidrule(lr){2-5} % 使用 \cmidrule 替代 \cline
& Bayer Model & 69.35 & 557.02 & - \\
& Ours & 68.82 & 80.44 & -85.36\% \\
\bottomrule
\end{tabular}%
}
\end{table}

\noindent \textbf{Runtime.} We measure the running time of our framework with PSPNet and Segformer on both CamVid and Cityscapes datasets, which are shown in Table~\ref{tab:runtime}. Our framework accelerates the back end video computer vision model in 4 to 8 times faster on a single NVIDIA A100 GPU. Specifically, our proposed ME processes $720\times960$ CamVid images in only 3.36 ms and $1024\times2048$ Cityscapes images in 4.44 ms.

\begin{table}
     \centering
  \small % <-- 1. 使用 \small 字体
  \setlength{\tabcolsep}{4pt} % <-- 2. 减少列间距 (默认为 6pt)
  \caption{Running time and FPS on NVIDIA A100.}
  \begin{tabular}{llcccc} % <-- 修改点：增加了一列 'l'
  \toprule
    % --- 修改后的表头 ---
  \multirow{2}{*}{\textbf{Model}} & \multirow{2}{*}{\textbf{Dataset}} & \multicolumn{2}{c}{\textbf{Frame-by-frame}} & \multicolumn{2}{c}{\textbf{Ours}} \\
  \cmidrule(lr){3-4} \cmidrule(lr){5-6} % <-- 修改点：列索引 +1
 & & Time(ms)  & FPS  & Time(ms)  & FPS \\
  \midrule
    % --- 修改后的数据行 ---
  \multirow{2}{*}{PSPNet} & CamVid & 47.42 & 21.09 & 14.02 & 71.33 \\
  & Cityscapes & 81.77 & 12.23 & 33.23 & 30.09 \\
  \midrule
  \multirow{2}{*}{Segformer} & CamVid & 136.82 & 7.31 & 21.47 & 46.58 \\
  & Cityscapes & 599.07 & 1.67 & 76.34 & 13.10 \\
  \bottomrule
  \end{tabular}
  \label{tab:runtime}
\end{table}

\begin{figure*}
    \centering
    \includegraphics[width=0.95\linewidth]{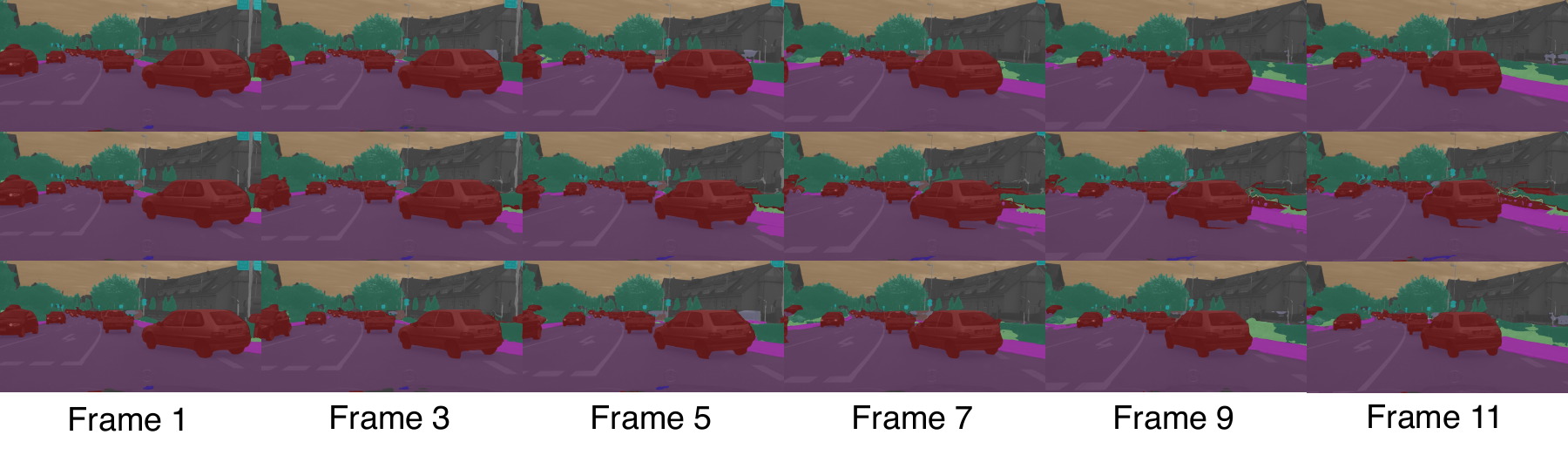}
    \caption{Visualization of our proposed method with PSPNet as the base model on video semantic segmentation task. First row: PSPNet frame-by-frame. Mid row: Prediction by motion vectors without perception residual. Last row: Results of our framework.}
    \label{fig:vis}
\end{figure*}

\noindent \textbf{Visualization}
The visualization of our framework is shown in Figure~\ref{fig:vis}. As illustrated, motion estimation effectively propagates perceptual information from the reference frame to the current frame. However, as estimation errors accumulate over time, noticeable artifacts may emerge. These artifacts are subsequently alleviated by the perception residual modules.

\begin{figure}[htbp]
	\centering
	\subfloat[Static Correction]{\includegraphics[width=.45\linewidth]{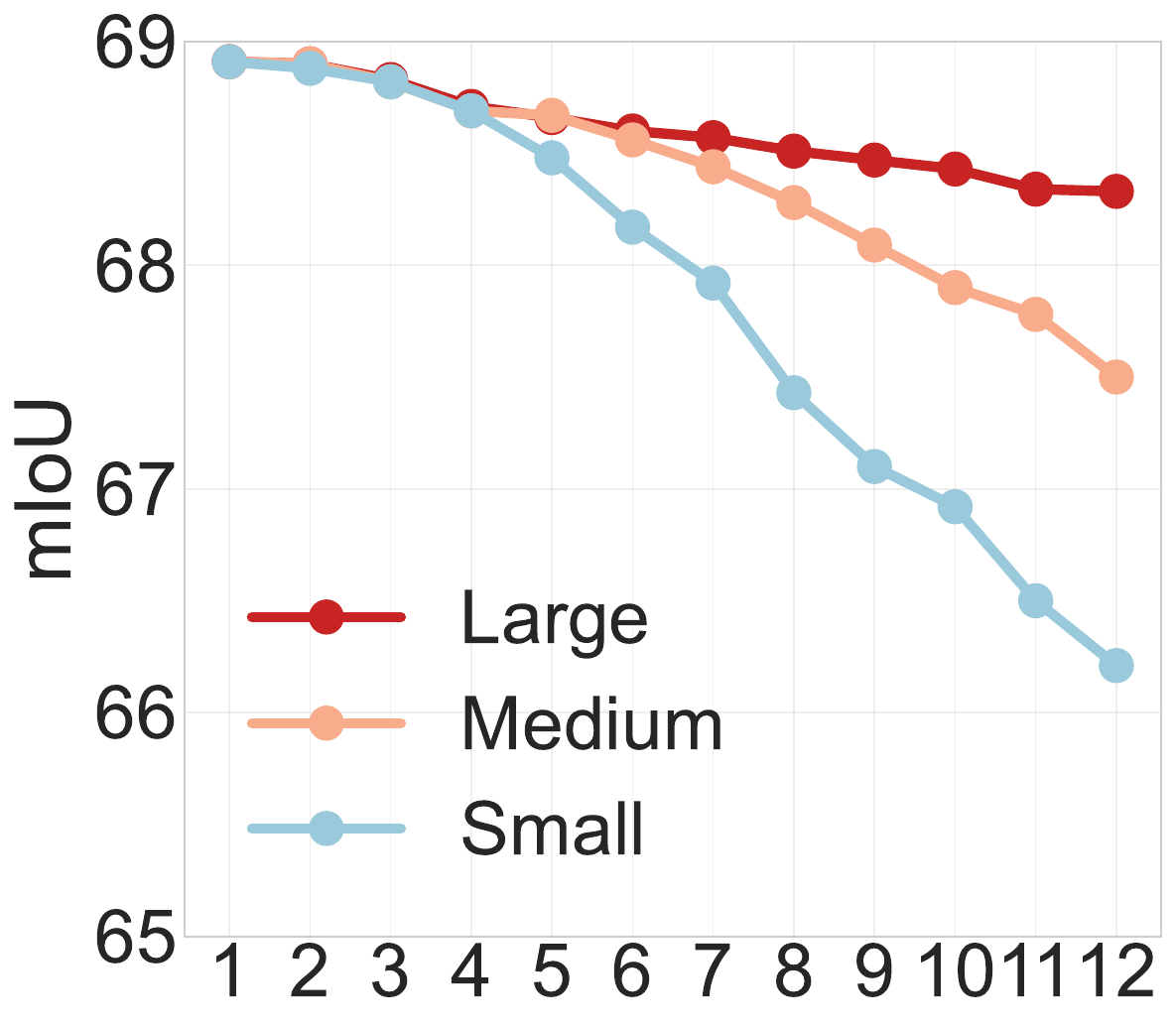}}\hspace{5pt}
	\subfloat[Dynamic Correction]{\includegraphics[width=.45\linewidth]{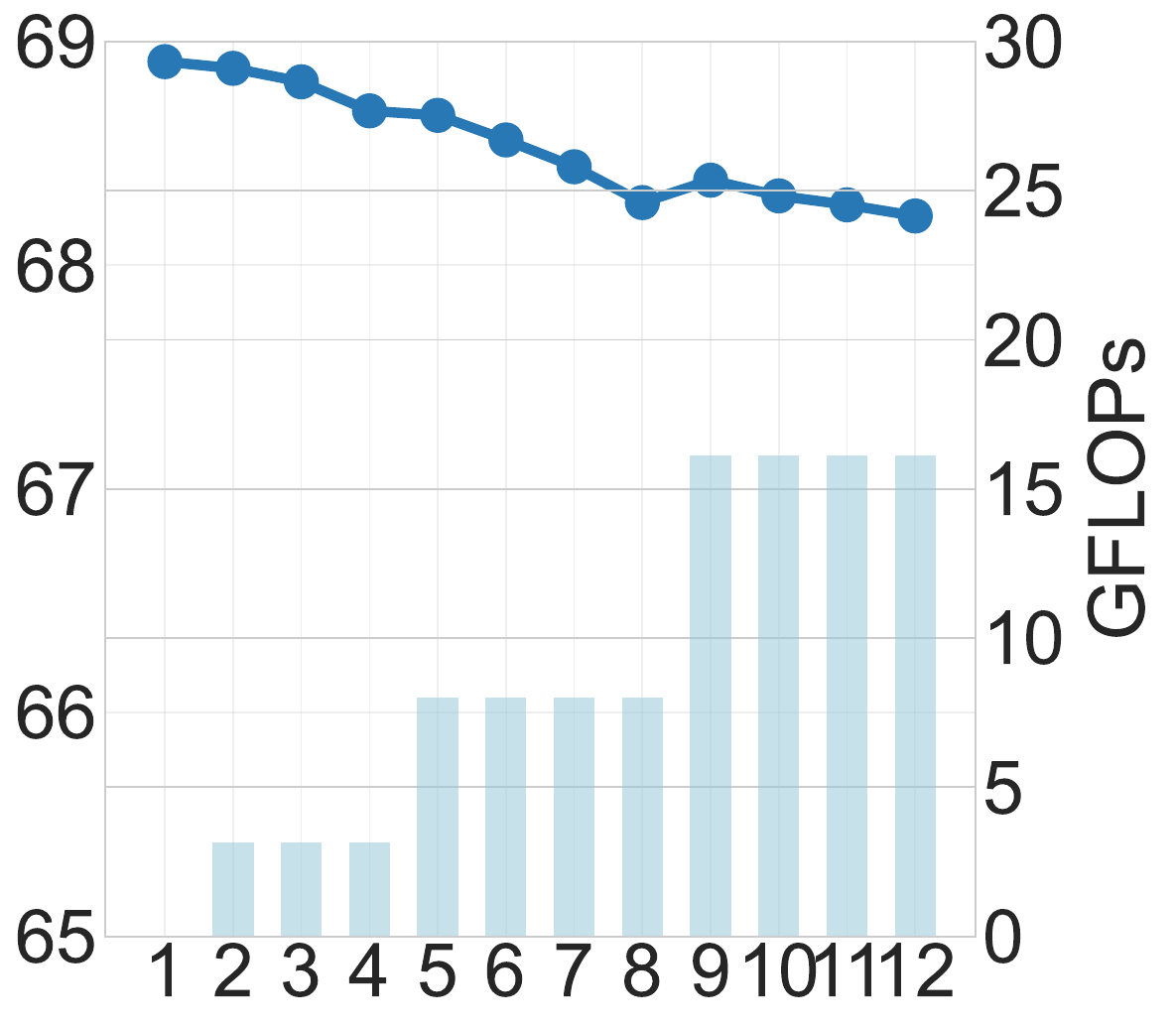}}\\
	\caption{Performance on Static and Dynamic Correction. Left: Performance using correction with 3 different scales. Right: Performance and FLOPs of dynamic correction.}
    \label{fig:dynamic}
\end{figure}

\noindent \textbf{Static and Dynamic Correction.} Correction can be performed using two strategies: employing models of a uniform scale or of varying scales. In a video sequence, the prediction error in the initial frames is typically smaller than in the later frames. The feature residual tends to become denser, primarily due to accumulated errors propagated from the previous frame or larger motion disparities when referencing the key frame. Therefore, to maintain stable performance, it is intuitive to use smaller models for sparser residuals and larger models for denser ones. As illustrated in Figure~\ref{fig:dynamic} (a), employing larger-scale correction models yields superior overall performance, particularly on later frames. For the initial frames, however, the performance remains comparable across different model scales. This observation motivates our test of a dynamic strategy: using small models for the initial frames and transitioning to larger models for the later ones, which is shown in Figure~\ref{fig:dynamic} (b). We allocate small models for the first three frames, medium-sized models for the middle four frames, and large models for the final four frames to better balance efficiency and accuracy across the sequence.

\begin{figure}[htbp]
	\centering
	\subfloat[Without Correction]{\includegraphics[width=.45\linewidth]{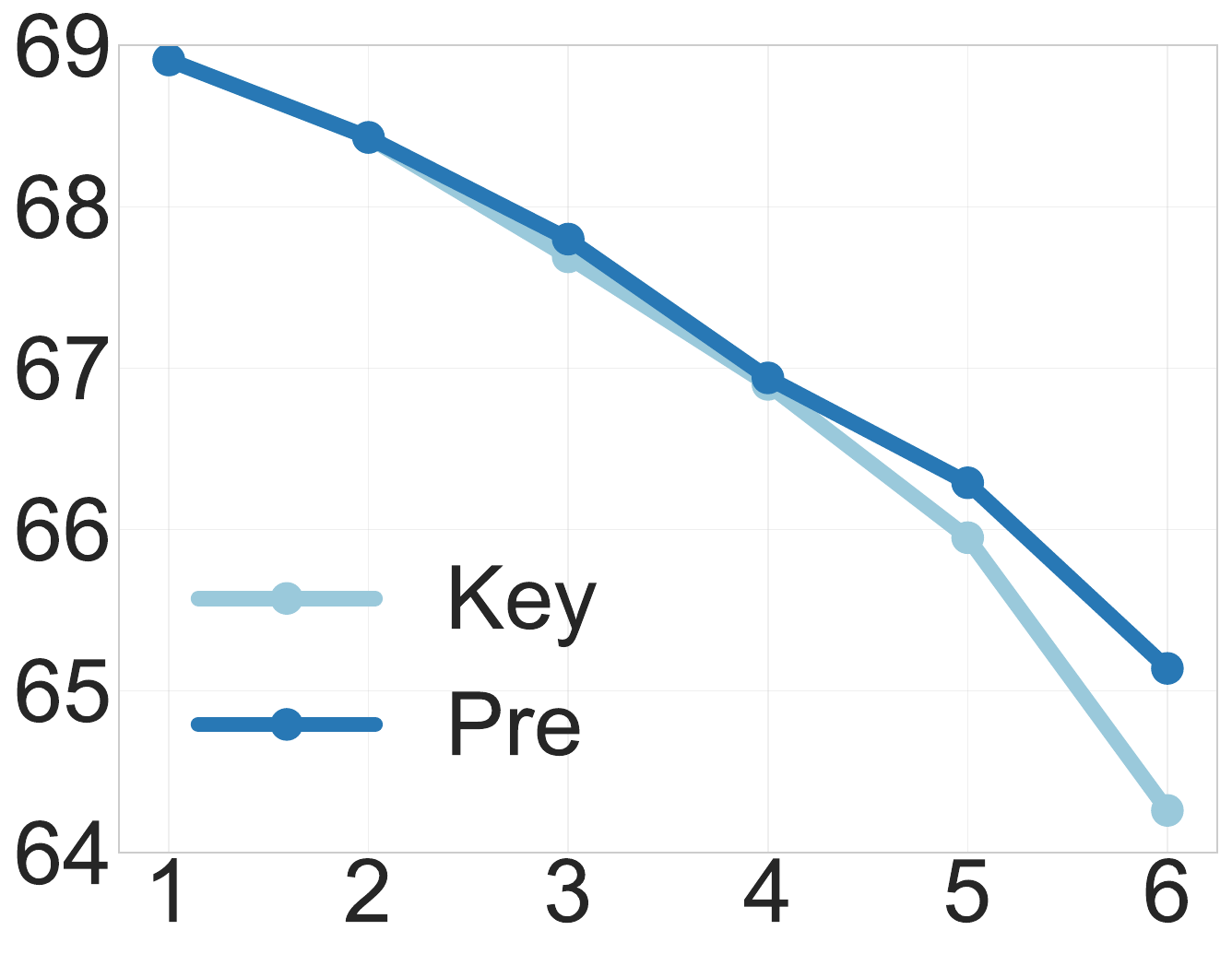}}\hspace{5pt}
	\subfloat[With Correction]{\includegraphics[width=.45\linewidth]{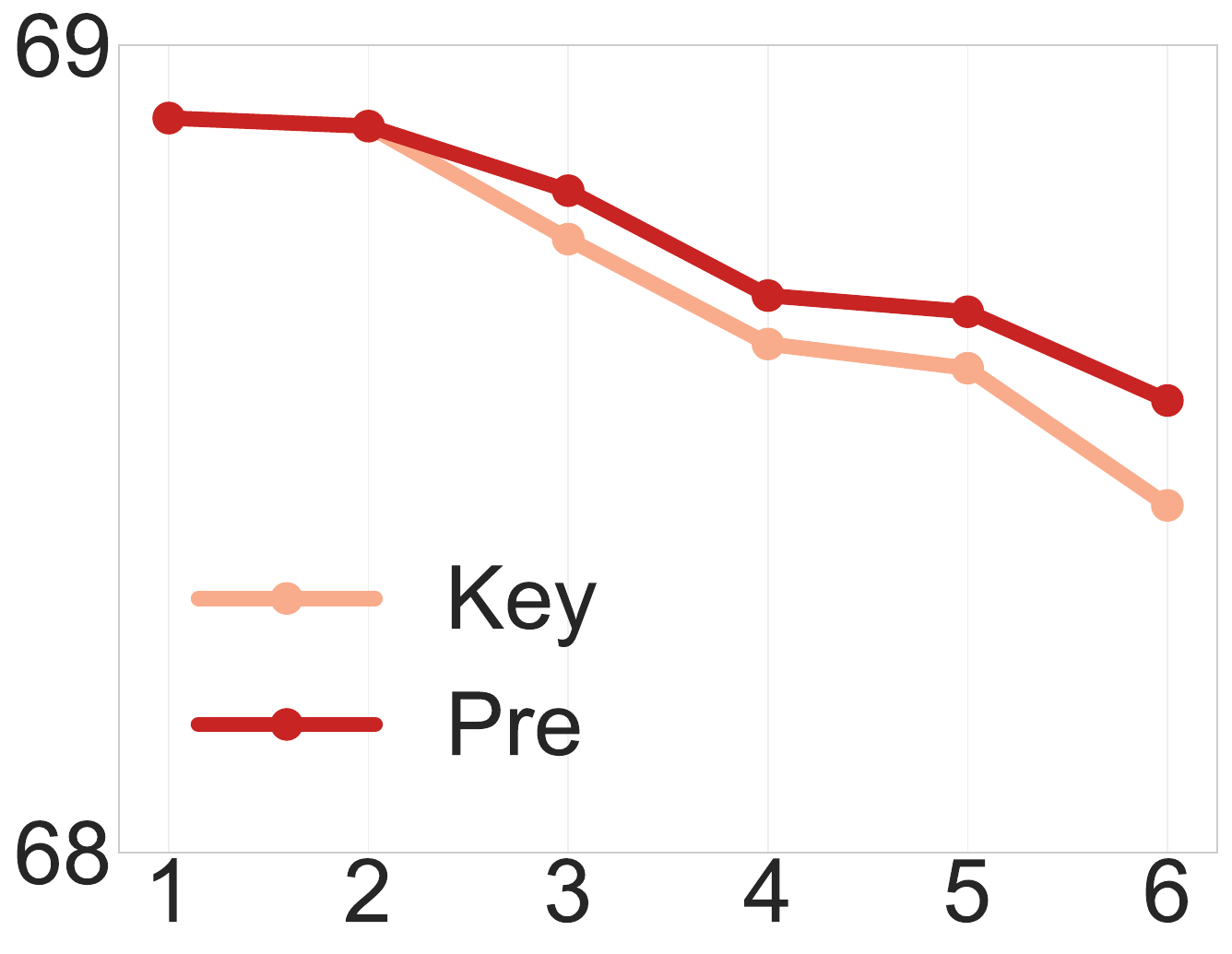}}\\
	\caption{Performance on different reference frames. Key/Pre means using the key/previous frame as the reference.}
    \label{fig:reference}
\end{figure}

\noindent \textbf{Performance on Different Reference Frames.} We test our framework using the key frame and the previous frame as reference respectively. As shown in Figure~\ref{fig:reference}, choosing the previous one frame as reference is usually better than choosing the key frame, because previous frame is more close to the current frame with smaller motion and better matching accuracy, both when not using correction and using static correction. 

% \noindent \textbf{GOP Length.} 
% Experiments on different GOP length are conducted, which is presented in Figure~\ref{fig:GOP}. Because the lightweight perception residual model for correction, with the increase of GOP lengths, the average FLOPs keeps decreasing, but with slight performance loss.

% \begin{figure}[htbp]
% 	\centering
% 	\subfloat[CamVid]{\includegraphics[width=.45\linewidth]{Figures/GOP_CamVid.pdf}}\hspace{5pt}
% 	\subfloat[Cityscapes]{\includegraphics[width=.45\linewidth]{Figures/GOP_Cityscapes.pdf}}\\
% 	\caption{Performance on different GOP lengths with PSPNet-18.}
%     \label{fig:GOP}
% \end{figure}

\begin{figure}
    \centering
    \includegraphics[width=0.8\linewidth]{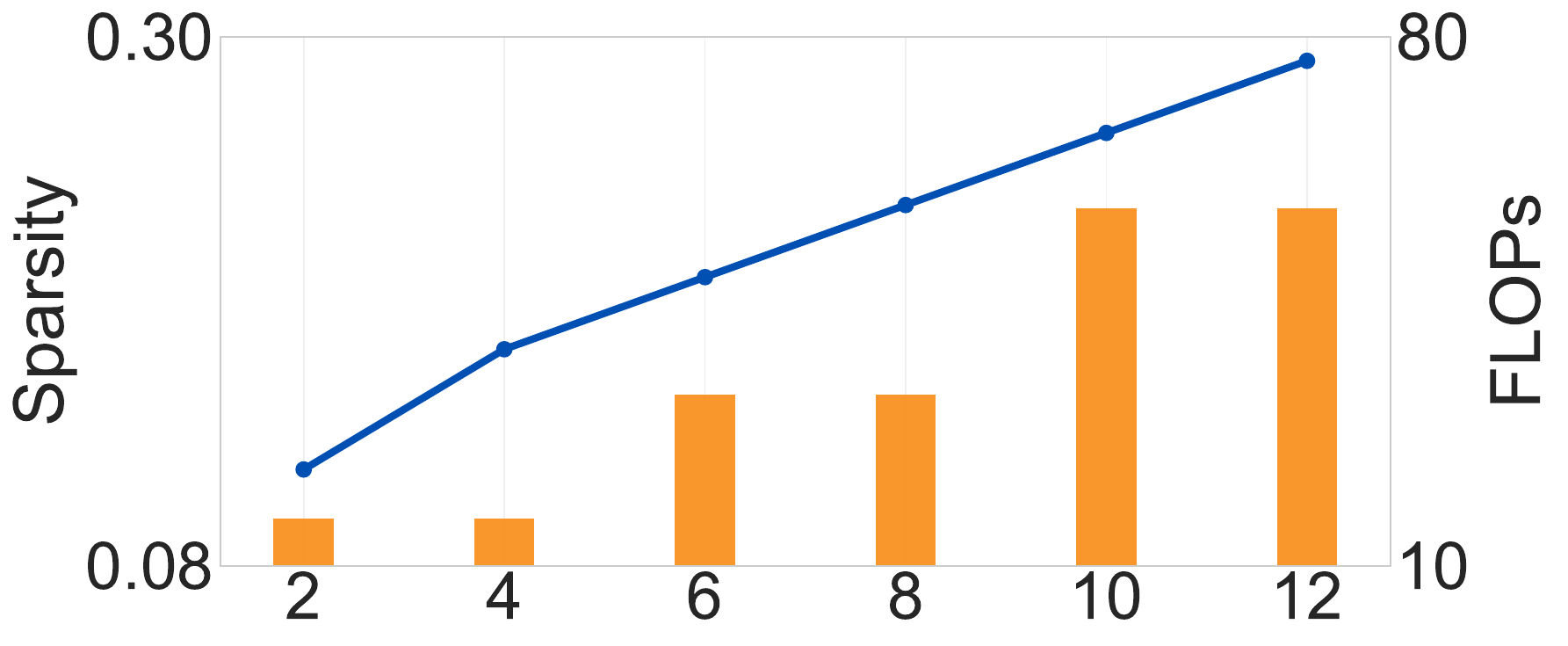}
    \caption{Relations of sparsity of perception residual and PRNet scale.}
    \label{fig:sparsity}
\end{figure}

\noindent \textbf{Sparsity of Perception Residual}
Analogous to video compression, a residual map with higher sparsity requires more storage. Similarly, in video computer vision, generating perception residuals with higher sparsity demands a larger model to preserve performance. As shown in Figure~\ref{fig:sparsity}, the residual sparsity increases with the frame index, and thus larger models are needed to maintain consistent performance.

\noindent \textbf{Training Loss.} To supervise the perception residual learning, as mentioned above, berHu loss is employed as the loss function, with region-size–based weighting. Other popular loss functions are also tested in our framework, as shown in Table~\ref{tab:loss}.
Loss functions with larger penalty for large errors exceeds the ones without it, proving the contribution of correction for large errors. BerHu Loss with weights outperforms the one without weights, proving the effectiveness and necessity of the region-size–based weighting.

\begin{table}[htbp]
    \centering
    \caption{Average mIoU with different loss functions on a GOP length of 12.}
    \begin{tabular}{ccc}
    \toprule
         & CamVid (\%) & Cityscapes (\%)\\
    \midrule
        weighted L1 & 67.79 & 67.26\\
        weighted L2 & 68.41 & 68.70\\
        weighted Huber & 67.94 & 68.02\\
        berHu & 66.45 & 65.83 \\
        weighted berHu & 68.53 & 68.82 \\
    \bottomrule
    \end{tabular}
    
    \label{tab:loss}
\end{table}

% \begin{table*}[h!]
%     \centering
%     \caption{Experiment results on PSPNet-18 under different FME settings without any correction.}
%     \label{tab:settings}

%     \begin{tabular}{cccccccccc}
%         \toprule
%         \multirow{2}{*}{Mode} & \multicolumn{2}{c}{Coarse} & \multicolumn{2}{c}{Intermediate} & \multicolumn{2}{c}{Fine}  & \multirow{2}{*}{Block Size} & \multirow{2}{*}{GFLOPs} & \multirow{2}{*}{mIoU (\%)} \\
%         \cmidrule(lr){2-3} \cmidrule(lr){4-5} \cmidrule(lr){6-7}
        
%         & Range & Step & Range & Step & Range & Step & & & \\
%         \midrule
%         Standard &  8 &  8 &  4 & 4  & 2  &  1  &  [64, 32] & 1.66 & 60.62 \\
%         \midrule
%         Mode 1 & 4 & 8 & 2 & 4 & 2 & 1  & [64, 32] & 0.55 &  60.17 \\
%         Mode 2 & 10  & 8  &  6 & 4  &  4 &  1  & [64, 32] & 2.90  &  59.89\\
%         \midrule
%         Mode 3 &  8 &  8 &  4 & 4  & 2  &  1  &  [32] & 0.83 &  60.34\\
%         Mode 4 &  8 &  8 &  4 & 4  & 2  &  1  &  [64,32,16,8] & 3.32 &  59.42 \\
%         \midrule
%         Mode 5 &  8 &  16 &  2 & 4  & 2  &  1  &  [64, 32] & 1.66 & 60.20 \\
%         Mode 6 &  8 &  4 &  2 & 4  & 2  &  1  &  [64, 32] & 1.66 &  60.55\\
%         \bottomrule
%     \end{tabular}
% \end{table*}

\subsection{Video Object Detection}
We also extended our framework's evaluation to video object detection. At a Group of Pictures (GOP) length of 12, our method achieves a 67.21\% reduction in FLOPs and an 82.29\% reduction in runtime. The acceleration in FLOPs is less pronounced than for video semantic segmentation, which we attribute to the lightweight design of the DETR base model. Notably, this significant speedup results in only a marginal mAP decrease, from 78.3\% to 77.9\%.

\begin{table}
    \centering
    \caption{Acceleration on video object detection.}
    \begin{tabular}{ccc}
    \toprule
         & DETR & Ours \\
         \midrule
        GFLOPs & 24.84 &  8.14\\
        Runtime(ms) & 27.23 &  4.82\\
        \bottomrule
    \end{tabular}
    
    \label{tab:detr}
\end{table}

\subsection{Comparison of Motion Estimation} 
To better illustrate the trade-off between accuracy and efficiency of our proposed fast ME, we compare it with representative motion estimation methods, as summarized in Table~\ref{tab:me}. The H.264 family can achieve relatively high speed under the ultrafast or fast presets, but its block-based matching is primarily designed for low-resolution videos and degrades significantly on high-resolution datasets such as Cityscapes. H.265 provides more accurate motion estimation due to its advanced prediction modes, yet the computational cost becomes prohibitively high, making it unsuitable for real-time scenarios.

Deep learning–based approaches, such as LiteFlowNet and RAFT-S, generally deliver strong accuracy. However, their inference latency remains far from meeting real-time requirements when operating on high-resolution inputs. In contrast, our proposed fast ME achieves a more favorable balance: it not only surpasses classical codecs in accuracy but also runs substantially faster than deep learning–based methods, demonstrating clear advantages for high-resolution and time-critical vision tasks.

\begin{table}
    \centering
    \caption{Comparison of different ME methods on Cityscapes with input resolution $1024\times 2048$.}
    \begin{tabular}{cccc}
    \toprule
        Type & Method & Runtime (ms) & mIoU\\
        \midrule
        \multirow{4}{*}{\rotatebox[origin=c]{90}{Codecs}}
         & H.264 Ultrafast & 15.5 & 57.92\\
         & H.264 Fast & 31.4 & 58.09\\
         & H.264 Slow & 52.7 & 59.13\\
         & H.265 & 201.02 & 61.02\\
         \midrule
         \multirow{2}{*}{\rotatebox[origin=c]{90}{DL}}& LiteFlowNet & 30.14 & 60.7\\
         & RAFT & 80.39 & 61.96\\
         \midrule
          & FME(standard) & 4.44 & 60.92 \\
         \bottomrule
    \end{tabular}
    
    \label{tab:me}
\end{table}

\subsection{Ablation Studies}
To demonstrate the effectiveness of each component in our efficient video computer vision inference, ablation studies are conducted with several variances:
\begin{enumerate}
    \item Frame-by-frame processing.
    \item With prediction by motion vectors, but without any correction.
    \item With prediction by motion vectors and perception residual correction.
\end{enumerate}
As presented in Table~\ref{tab:ablation}, the naive frame-by-frame scheme incurs the highest computational cost due to massive redundancies across frames. Utilizing only motion vector prediction, driven by our fast motion estimation, significantly reduces this computational load. However, this efficiency comes at the cost of notable performance degradation. The introduction of our perception residual model effectively corrects this deficit, restoring high-level performance. This demonstrates our full model's ability to achieve a favorable trade-off, maintaining high accuracy with limited computational overhead.

\begin{table}[htbp]
    \centering
    \caption{Ablation studies with a GOP length of 12.}
    \begin{tabular}{ccc}
    \toprule
        Variance & mIoU (\%) & GFLOPs\\
    \midrule
        1 & 69.35 & 557.02\\
        2 & 60.62 & 47.94\\
        3 & 68.82 & 80.44\\
    \bottomrule
    \end{tabular}
    
    \label{tab:ablation}
\end{table}

\section{Conclusion}
In this paper, we propose an efficient Bayer-domain video computer vision system, solving two fundamental problems:  1) how to eliminate unnecessary computation from the pipeline, and 2) how to effectively reduce temporal redundancy while preserving accuracy with minimal additional computational overhead. On the front end, we remove the image signal processor from the traditional RGB-domain pipeline and directly process Bayer raw data with back end models, saving the computation and energy costs. On the back end, we introduce motion estimation to extract the temporal correlation across the frames, thus avoiding redundant computation on similar contents. To repair the artifacts from mismatching, we propose perception residual, extending the residual map into perception level and numerically correcting the prediction with lightweight models. Experiments are conducted on video semantic segmentation and video object detection tasks, achieving significant acceleration with slight performance loss.
{
    \small
    \bibliographystyle{ieeenat_fullname}
    \bibliography{main}
}


\end{document}